\documentclass[10pt,twocolumn,letterpaper]{article}

\usepackage{iccv}
\usepackage{times}
\usepackage{epsfig}
\usepackage{graphicx}
\usepackage{amsmath}
\usepackage{amssymb}

\usepackage[ruled]{algorithm2e}
\usepackage{multirow}
\usepackage{array}
\usepackage{ctable} % for \specialrule command
\usepackage{mathptmx}
\DeclareMathAlphabet{\mathcal}{OMS}{cmsy}{m}{n}
\usepackage{textcomp}
\usepackage{dblfloatfix}
\usepackage{bm}

\newcommand{\KG}{\textcolor{black}}
\newcommand{\cc}{\textcolor{black}}

\newcommand{\RG}{\textcolor{black}}

\DeclareMathOperator*{\argmin}{\arg\!\min}
\newcolumntype{?}[1]{!{\vrule width #1}}

\usepackage[pagebackref=true,breaklinks=true,letterpaper=true,colorlinks=false,bookmarks=true]{hyperref}

\iccvfinalcopy % *** Uncomment this line for the final submission

\def\iccvPaperID{555} % *** Enter the ICCV Paper ID here

\begin{document}

%%%%%%%%% TITLE
\title{On-Demand Learning for Deep Image Restoration}

\author{Ruohan Gao and Kristen Grauman\\
University of Texas at Austin\\
{\tt\small \{rhgao,grauman\}@cs.utexas.edu}
}

\maketitle
%\thispagestyle{empty}

%===========================================================
%%%%%%%%% ABSTRACT
\begin{abstract}
While machine learning approaches to image restoration offer great promise, current methods risk training models fixated on performing well only for image corruption of a particular level of difficulty---such as a certain level of noise or blur. First, we examine the weakness of conventional ``fixated" models and demonstrate that training general models to handle arbitrary levels of corruption is indeed non-trivial. Then, we propose an \emph{on-demand} learning algorithm for training image restoration models with deep convolutional neural networks. The main idea is to exploit a feedback mechanism to self-generate training instances where they are needed most, thereby learning models that can generalize across difficulty levels.  On four restoration tasks---image inpainting, pixel interpolation, image deblurring, and image denoising---and three diverse datasets, our approach consistently outperforms both the status quo training procedure and curriculum learning alternatives.  
\vspace{-12pt}
\end{abstract}
%===========================================================
%%%%%%%%% INTRODUCTION
\vspace{-0.1in}
\section{Introduction}
\label{sec:intro}

Deep convolutional networks~\cite{krizhevsky2012imagenet,simonyan2014very,he2015deep} have swept the field of computer vision and have produced stellar results on various recognition benchmarks in the past several years. Recently, deep learning methods are also becoming a popular choice to solve low-level vision tasks in image restoration, with exciting results~\cite{dong2016image,liu2016learning,larsson2016learning,zhang2016colorful,cho2016natural,johnson2016perceptual,pathak2016context,yeh2016semantic}. Restoration tasks such as image super-resolution, inpainting,  deconvolution,  matting,  and colorization have a wide range of compelling applications. \RG{For example, deblurring techniques can mitigate motion blur in photos, and denoising methods can recover images corrupted by sensor noise.}

\begin{figure}
  \centering
  \includegraphics[scale=0.93]{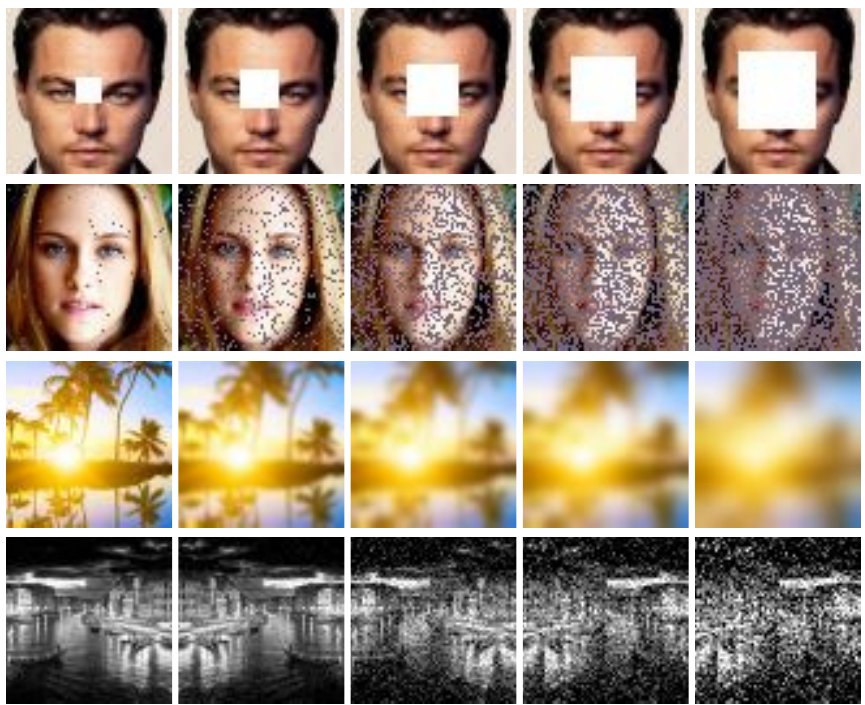}
  \caption{Illustration of four image restoration tasks: image inpainting, pixel interpolation, image deblurring, and image denoising. Each task exhibits increasing difficulty based on size of inpainting area, percentage of deleted pixels, degree of blurriness, and severity of noise. Our work aims to train all-rounder models that perform well across the spectrum of difficulty for each task. }
  \label{fig:concept}
  \vspace{-12pt}
\end{figure}

A learning-based approach to image restoration enjoys the convenience of being able to \emph{self-generate} training instances purely based on the original real images.  Whereas training an object recognition system entails collecting images manually labeled with object categories by human annotators, an image restoration system can be trained with arbitrary, synthetically corrupted images. The original image itself is the ground-truth the system learns to recover.

While existing methods take advantage of this convenience, they typically do so in a problematic way.  Image corruption exists in various degrees of severity, and so in real-world applications the difficulty of restoring images will also vary significantly. For example, as shown in Fig.~\ref{fig:concept}, an inpainter may face images with varying sizes of missing content, and a deblurring system may encounter varying levels of blur.  Intuitively, the more missing pixels or the more severe the blur, the more difficult the restoration task.

However, the norm in existing deep learning methods is to train a model that succeeds at restoring images exhibiting \emph{a particular level of corruption difficulty}.  In particular, existing systems self-generate training instances with a
manually fixed hyper-parameter that controls the degree of corruption---a fixed inpainting size~\cite{pathak2016context,yeh2016semantic}, a fixed percentage of corrupted pixels~\cite{yeh2016semantic,liu2016learning}, or a fixed level of white Gaussian noise~\cite{liu2016learning,xie2012image,jain2009natural,burger2012image}.  
The implicit assumption is that at test time, either i) corruption will be limited to that same difficulty, or ii) some other process, e.g.,~\cite{liu2006noise,liu2013single,chen2015efficient}, will estimate the difficulty level before passing the image to the appropriate, separately trained restoration system.  
Unfortunately, these are strong assumptions that remain difficult to meet in practice.  As a result, existing methods risk training \RG{\emph{fixated models}: models that perform well only at a particular level of difficulty}. Indeed, deep networks can severely overfit to a certain degree of corruption. Taking the inpainting task as an example, a well-trained deep network may be able to inpaint a $32 \times 32$ block \cc{out of a $64 \times 64$ image} very well, then fails miserably at inpainting a (seemingly easier) $10 \times 10$ block (see Fig.~\ref{fig:overfitting} and Sec.~\ref{sec:overfitting_examples}). Furthermore, as we will show, simply pooling training instances across all difficulty levels makes the deep network struggle to adequately learn the concept.

\label{intro:analogy}
How should we train an image restoration system to succeed across a spectrum of difficulty levels? In this work we explore ways to let a deep learning system take control and guide its own training. This includes i) a solution that simply pools training instances from across difficulty levels, ii) a solution that focuses on easy/hard examples, iii) curriculum learning solutions that intelligently order the training samples from easy to hard, and iv) a new \emph{on-demand learning} solution for training general deep networks across difficulty levels. Our approach relies on a feedback mechanism that, at each epoch of training, lets the system guide its own learning towards the right proportion of sub-tasks per difficulty level. In this way, the system itself can discover which sub-tasks deserve more or less attention.    

To implement our idea, we devise a general encoder-decoder network amenable to several restoration tasks.  We evaluate the approach on four low-level tasks---inpainting, pixel interpolation, image deblurring, and denoising---and three diverse datasets, CelebFaces Attributes~\cite{liu2015faceattributes}, SUN397 Scenes~\cite{xiao2014sun}, and the Denoising Benchmark 11 (DB11)~\cite{dabov2007image,burger2012image}.  Across all tasks and datasets, the results consistently demonstrate the advantage of our proposed method.  On-demand learning helps avoid the common (but thus far neglected) pitfall of overly specializing deep networks to a narrow band of distortion difficulty.
%===========================================================
%%%%%%%%% RELATED WORK
\vspace{-0.1in}
\section{Related Work}
\paragraph{\textbf{Deep Learning in Low-Level Vision:}}
Deep learning for image restoration is on the rise.  Vincent~\etal~\cite{vincent2008extracting} propose one of the most well-known models: the stacked denoising auto-encoder. A multi-layer perceptron (MLP) is applied to image denoising by Burger~\etal~\cite{burger2012image} and post-deblurring denoising by~Schuler~\etal~\cite{schuler2013machine}. Convolutional neural networks are also applied to natural image denoising~\cite{jain2009natural} and used to remove noisy patterns (e.g., dirt/rain)~\cite{eigen2013restoring}. Apart from denoising, deep learning is gaining traction for various other low-level tasks: super-resolution~\cite{dong2016image,johnson2016perceptual}, inpainting~\cite{pathak2016context,yeh2016semantic}, deconvolution~\cite{xu2014deep}, matting~\cite{cho2016natural}, and colorization~\cite{larsson2016learning,zhang2016colorful}. While many models specialize the architecture towards one restoration task, recent work by Liu~\etal presents a unified network for multiple tasks~\cite{liu2016learning}. Our encoder-decoder pipeline also applies across tasks, and serves as a good testbed for our main contribution---the idea of on-demand learning. Our idea has the potential to benefit any existing method currently limited to training with a narrow band of difficulty~\cite{pathak2016context,yeh2016semantic,jain2009natural,burger2012image,schuler2013machine,liu2016learning}.

\RG{The fixation problem} is also observed in recent denoising work, e.g., \cite{burger2012image,mao2016image}, but without a dedicated and general solution.  Burger~\etal~\cite{burger2012image} attempt to train a network on patches corrupted by noise with different noise levels by giving the noise hyper-parameter as an additional input to the network. While the model can better denoise images at different noise levels, \KG{assuming the noise level is known at test time is problematic.} Recently, Mao~\etal~\cite{mao2016image} explore how the large capacity of a very deep network can help generalize across noise levels, but accuracy still declines noticeably from the fixated counterpart.

\vspace{-0.2in}

\paragraph{\textbf{Curriculum and Self-Paced Learning:}}
\RG{Training neural networks according to a \emph{curriculum} can be traced back at least to Elman~\cite{elman1993learning}. Prior work mainly focuses on supervised learning and a single task, like the seminal work of Bengio~\etal \cite{bengio2009curriculum}. Recently, Pentina~\etal~\cite{pentina2015curriculum} pose curriculum learning in a multi-task learning setting, where sharing occurs only between subsequent tasks.
Building on the curriculum concept, in \emph{self-paced} learning,  the system \emph{automatically} chooses the order in which training examples are processed~\cite{kumar2010self,lee2011learning}. We are not aware of any prior work in curriculum/self-paced learning that deals with image restoration.  Like self-paced learning, our approach does not rely on human annotations to rank training examples from easiest to hardest. Unlike self-paced work, however, our on-demand approach self-generates training instances of a targeted difficulty.}

\vspace{-0.2in}

\paragraph{\textbf{Active Learning:}}
Active learning is another way for a learner to steer its own learning. Active learning selects examples that seem most valuable for human labeling, and has been widely used in computer vision to mitigate manual annotation costs~\cite{kapoor2010gaussian,huang2010active,elhamifar2013convex,vijayanarasimhan2014large,li2014multi,freytag2014selecting,kading2015active,wang2016multi}. Unlike active learning, our approach uses no human annotation, but instead actively synthesizes training instances of different corruption levels based on the progress of training. \cc{All our training data can be obtained for ``free" and the ground-truth (original uncorrupted image) is always available.}
%===========================================================
%%%%%%%%% APPROACH
\vspace{-0.05in}
\section{Roadmap}
\vspace{-0.05in}
We first examine \RG{the fixation problem}, and provide concrete evidence that it hinders deep learning for image restoration (Sec.~\ref{sec:overfitting_examples}). Then we present a unified view of image restoration as a learning problem (Sec.~\ref{sec:formulation}) and describe inpainting, interpolation, deblurring, and denoising as instantiations (Sec.~\ref{sec:tasks}). Next we introduce the on-demand learning idea (Sec.~\ref{sec:ondemand}) and our network architecture (Sec.~\ref{sec:network}). Finally, we present results (Sec.~\ref{sec:expts}).

\vspace{-0.05in}
\section{The Fixation Problem}
\vspace{-0.05in}
\label{sec:overfitting_examples}

\begin{figure}
  \centering
  \includegraphics[scale=0.48]{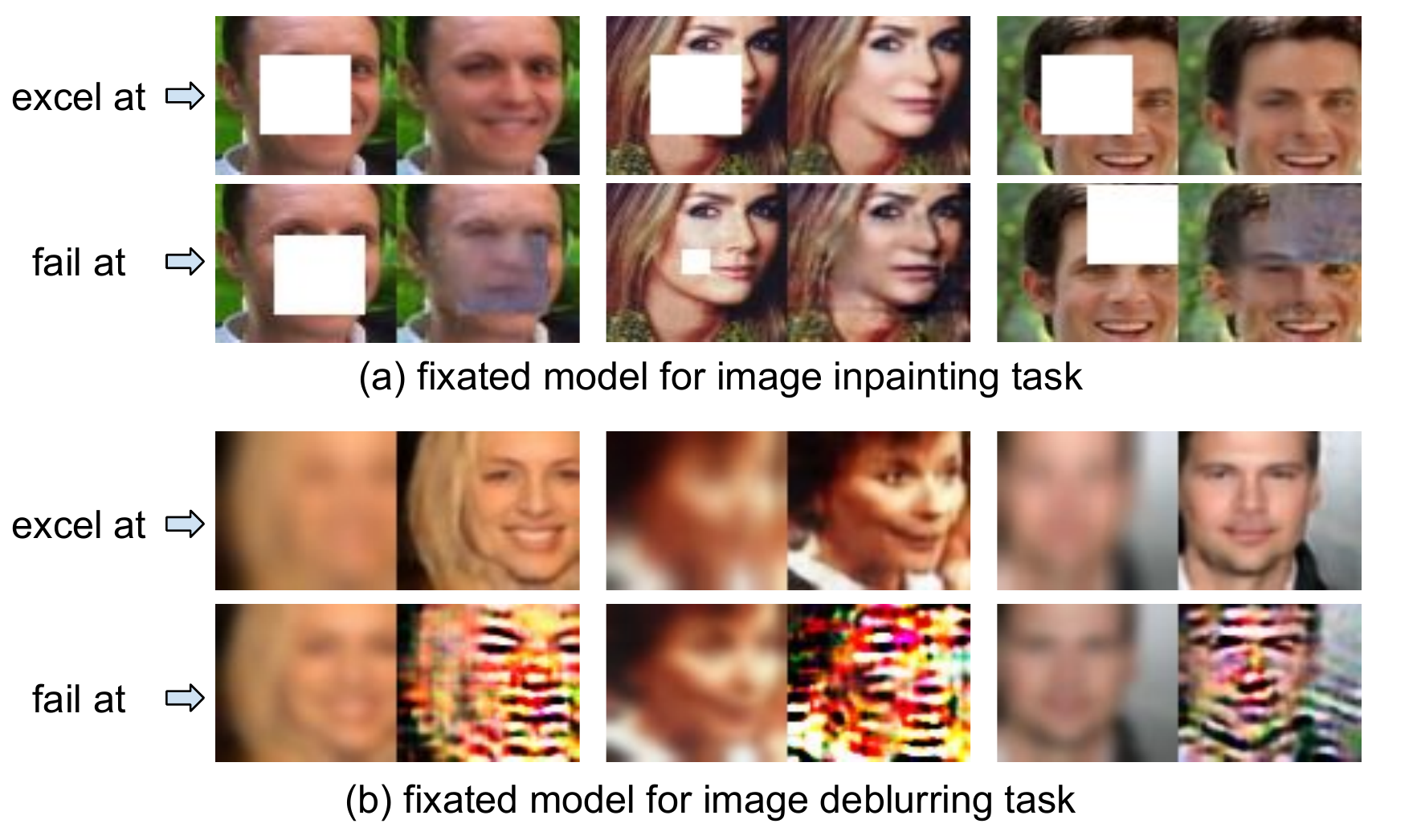}
  \caption{Illustration of the severity of overfitting for image inpainting and deblurring. The models overfit to a certain degree of corruption. They perform extremely well at that level of corruption, yet fail to produce satisfactory restoration results even for much easier sub-tasks. See Supp.~for other tasks and details.}
  \label{fig:overfitting}
  \vspace{-12pt}

\end{figure}

\RG{The fixation problem} arises when existing image restoration methods train a learning algorithm to restore images with a controlled degree of corruption~\cite{xie2012image,yeh2016semantic,burger2012image,schuler2013machine,pathak2016context,liu2016learning}.
For example, Yeh~\etal~\cite{yeh2016semantic} train an image inpainter at a fixed size and location, and always delete 80\% of pixels for pixel interpolation.  Pathak~\etal~\cite{pathak2016context} mainly focus on a large central block for the inpainting task. Liu~\etal~\cite{liu2016learning} solve denoising, pixel interpolation, and color interpolation tasks all with a restricted degree of corruption. While such methods may fix the level of corruption in training as a proof of concept, they nonetheless do not offer a solution to make the model generally applicable.

Just how bad is \RG{the fixation problem} in image restoration tasks?  Fig.~\ref{fig:overfitting} helps illustrate. To get these results, we followed the current literature to train deep networks to target a certain degree of corruption for four applications (See Supp.~for similar results of interpolation and denoising).\footnote{See Sec.~\ref{sec:expts} for quantitative results, and Sec.~\ref{sec:network} for details about the encoder-decoder network used.}

Specifically, for the image inpainting task, following similar settings of prior work~\cite{pathak2016context,yeh2016semantic}, we train a model to inpaint a large central missing block of size $32 \times 32$. During testing, the resulting model can inpaint the central block of the same size at the same location very well (first row in Fig.~\ref{fig:overfitting}-a).
    However, if we remove a block that is slightly shifted away from the central region, or remove a much \emph{smaller} block, the model fails to inpaint satisfactorily (second row in Fig.~\ref{fig:overfitting}-a). \KG{For the deblurring results in Fig.~\ref{fig:overfitting} (and interpolation \& denoising results in Supp.), we attempt analogous trials, i.e., training for 80\% missing pixels~\cite{yeh2016semantic}, a single width blur kernel or a single noise level, respectively, then observe poor performance by the fixated model on examples having different corruption levels.}

The details of the deep networks used to generate the results in Fig.~\ref{fig:overfitting} are not identical to those in prior work. However, we stress that the limitation in their design that we wish to highlight is orthogonal to the particular architecture. To apply them satisfactorily in a general manner would require training a separate model for each hyper-parameter. Even if one could do so, it is difficult to gauge the corruption level in a novel image and decide which model to use. Finally, as we will see below, simply pooling training instances across all difficulty levels is also inadequate.

\vspace{-0.05in}
\section{Approach}
\vspace{-0.05in}
Next we present ideas to overcome the fixation problem.
\vspace{-0.05in}

\subsection{Problem Formulation}\label{sec:formulation}
\vspace{-0.05in}

While the problem of overfitting is certainly not limited to image restoration, both the issue we have exposed as well as our proposed solution are driven by its special ability to self-generate ``free" training instances under specified corruption parameters.  Recall that a  real training image automatically serves as the ground-truth; the corrupted image is synthesized by applying a randomized corruption function.

We denote a real image as $\mathcal{R}$ and a corrupted image as $\mathcal{C}$ (e.g., a random block is missing). We model their joint probability distribution by $p(\mathcal{R},\mathcal{C}) = p(\mathcal{R})p(\mathcal{C}|\mathcal{R})$, where $p(\mathcal{R})$ is the distribution of real images and $p(\mathcal{C}|\mathcal{R})$ is the distribution of corrupted images given the original real image. In the case of a fixated model, $\mathcal{C}$ may be a deterministic function of $\mathcal{R}$ (e.g., specific blur kernel). 

To restore the corrupted image, the most direct way is to find $p(\mathcal{R}|\mathcal{C})$ by applying Bayes' theorem. However, this is not feasible because $p(\mathcal{R})$ is intractable. Therefore, we resort to a point estimate $f(\mathcal{C}, \textbf{w})$ through an encoder-decoder style deep network \textbf{(details in Sec.~\ref{sec:network})} by minimizing the following mean squared error objective:
\vspace{-0.05in}
\begin{equation} \label{eq:1}
 	\mathbb{E}_{\mathcal{R},\mathcal{C}} ||\mathcal{R} - f(\mathcal{C}, \textbf{w})||_2^2.
\vspace{-0.05in}
\end{equation}

Given a corrupted image $\mathcal{C}_0$, the minimizer of the above objective is the conditional expectation: $\mathbb{E}_{\mathcal{R}}[\mathcal{R}|\mathcal{C}=\mathcal{C}_0]$, which is the average of all possible real images that could have produced the given corrupted image $\mathcal{C}_0$.

Denote the set of real images $\{\mathcal{R}_i\}$.  We synthesize corrupted images $\{\mathcal{C}_i\}$ correspondingly to produce training image pairs $\{\mathcal{R}_i,\mathcal{C}_i\}$. We train our deep network to learn its weights $\textbf{w}$ by minimizing the following Monte-Carlo estimate of the mean squared error objective:

\vspace{-0.05in}
\begin{equation}
	\hat{\textbf{w}} = \argmin_{\textbf{w}} \sum_{i}||\mathcal{R}_i - f(\mathcal{C}_i, \textbf{w})||_2^2.
\vspace{-0.05in}
\end{equation}
During testing, our trained deep network takes a corrupted image $\mathcal{C}$ as input and forwards it through the network to output $f(\mathcal{C},\textbf{w})$ as the restored image.

\subsection{Image Restoration Task Descriptions}\label{sec:tasks}
Under this umbrella of a general image restoration solution, we consider four tasks.

\vspace*{-0.18in}
\paragraph{Image Inpainting}

The image inpainting task aims to refill a missing region and reconstruct the real image $\mathcal{R}$ of an incomplete corrupted image $\mathcal{C}$ (e.g., with a contiguous set of pixels removed). In applications, the ``cut out" part of the image would represent an occlusion, cracks in photographs, or an object that should be removed from the photo. Unlike~\cite{pathak2016context,yeh2016semantic}, we make the missing square block randomized across the whole image in both position and scale.

\vspace*{-0.18in}

\paragraph{Pixel Interpolation}
Related to image inpainting, pixel interpolation aims to refill non-contiguous deleted pixels. The network has to reason about the image structure and infer values of the deleted pixels by interpolating from neighboring pixels. Applications include more fine-grained inpainting tasks such as removing dust spots in film.

\vspace*{-0.18in}
\paragraph{Image Deblurring}

The image deblurring task aims to remove the blurring effects of a corrupted image $\mathcal{C}$ to restore the corresponding real image $\mathcal{R}$. We use Gaussian smoothing to blur a real image to create training examples. The kernel's horizontal and vertical widths ($\sigma_x$ and $\sigma_y$) control the degree of blurriness and hence the difficulty.  \KG{Applications include removing motion blur or defocus aberration.}

\vspace*{-0.18in}
\paragraph{Image Denoising}
The image denoising task aims to remove additive white Gaussian (AWG) noise of a corrupted image $\mathcal{C}$ to restore the corresponding real image $\mathcal{R}$. We corrupt real images by adding noise drawn from a zero-mean normal distribution with variance $\sigma$ (the noise level). 

\subsection{On-Demand Learning for Image Restoration}\label{sec:ondemand}

All four image restoration tasks offer a spectrum of difficulty. The larger the region to inpaint, the larger the percentage of deleted pixels, the more blurry the corrupted image, or larger the variance of the noise, the more difficult the corresponding task. To train a system that generalizes across task difficulty, a natural approach is to simply pool training instances across all levels of difficulty, insisting that the learner simultaneously tackle all degrees of corruption at once. Unfortunately, as we will see in our experiments, this approach can struggle to adequately learn the concept.

Instead, we present an \emph{on-demand} learning approach in which the system dynamically adjusts its focus where it is most needed.  First, we divide each restoration task into $N$ sub-tasks of increasing difficulty.  During training, we aim to jointly train the deep neural network restoration model (architecture details below) to accommodate all $N$ sub-tasks.  Initially, we \KG{generate} the same number of training examples from each sub-task in every batch. At the end of every epoch, we validate on a small validation set and evaluate the performance of the current model on all sub-tasks. \RG{We compute the mean peak signal-to-noise ratio (PSNR) for all images in the validation set for each sub-task.\footnote{PSNR is widely used as a good approximation to human perception of quality in image restoration tasks. We found PSNR to be superior to an L2 loss; because it is normalized by the max possible power and expressed in log scale, it is better than L2 at comparing across difficulty levels.}} A lower PSNR indicates a more difficult sub-task, suggesting that the model needs more training on examples of this sub-task. Therefore, we \KG{generate more} training examples for this sub-task in each batch in the next epoch. \KG{That is, we re-distribute the corruption levels allocated to the same set of training images.}  Specifically, we assign training examples in each batch for the next epoch inversely proportionally to the mean PSNR $P_i$ of each sub-task $T_i$. Namely,
\vspace*{-0.05in}
\begin{equation}
B_i = \frac{1/P_i}{\sum_{i=1}^{N}1/P_i} \cdot \mathbb{B},
\vspace*{-0.05in}
\end{equation}
where $\mathbb{B}$ is the batch size and $B_i$ is the number of of training examples assigned to sub-task $T_i$ for the next epoch. Please see Supp.~for the pseudocode of our algorithm.

On-demand learning bears some resemblance to boosting and hard negative mining, in that the system refocuses its effort on examples that were handled unsatisfactorily by the model in previous iterations of learning. However, whereas they reweight the influence given to individual (static) training samples, our idea is to self-generate \emph{new} training instances in specified difficulty levels based on the model's current performance. Moreover, the key is not simply generating more difficult samples, but to let the network steer its own training process, and decide how to schedule the \emph{right proportions} of difficulty.

Our approach discretizes the difficulty space via its intrinsic continuity property for all tasks. However, it is the network itself that determines the difficulty level for each discretized bin based on the restoration quality (PSNR) from our algorithm, and steers its own training.

We arrived at this simple but effective approach after investigating several other schemes inspired by curriculum and multi-task learning, as we shall see below. In particular, we also developed a new curriculum approach that stages the training samples in order of their difficulty, starting with easier instances (less blur, smaller cut-outs) \cc{for the system to gain a basic representation}, then moving onto harder ones (more blur, bigger cut-outs).  \cc{Wary that what appears intuitively easier to us as algorithm designers need not be easier to the deep network, we also considered an ``anti-curriculum" approach that reverses that ordering, e.g., starting with bigger missing regions for inpainting.} More details are given in Sec.~\ref{baselines}. 

\vspace*{-0.05in}
\subsection{Deep Learning Network Architecture}
\vspace*{-0.05in}

\label{sec:network}
\begin{figure}
  \centering
  \includegraphics[scale=0.75]{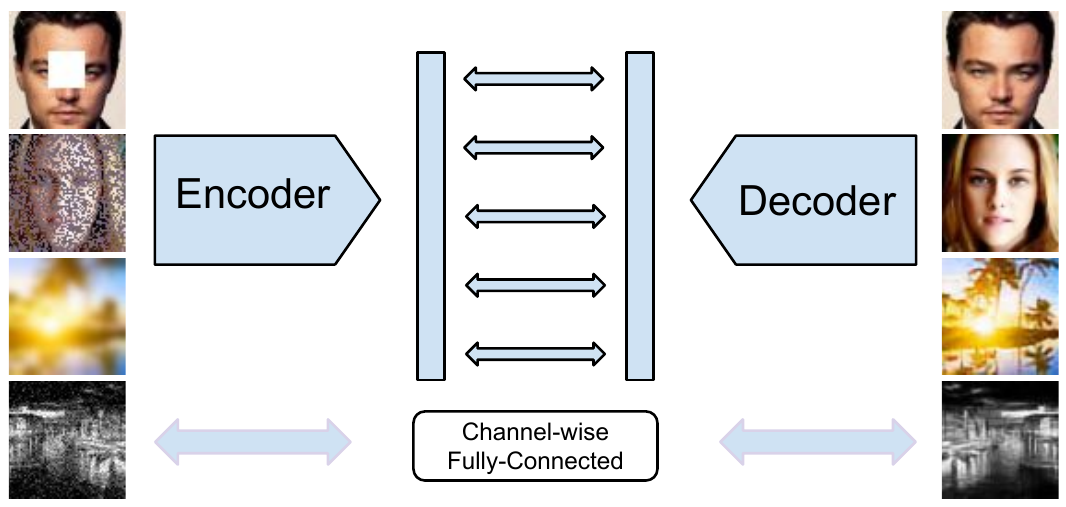}
  \caption{Network architecture for our image restoration framework, an encoder-decoder pipeline connected by a channel-wise fully-connected layer. See Supp. for details.}
  \label{fig:network}
  \vspace*{-0.18in}
\end{figure}

Finally, we present the network architecture used for all tasks to implement our on-demand learning idea.  Our image restoration network is a simple encoder-decoder pipeline.  See Fig.~\ref{fig:network}.  The encoder takes a corrupted image $\mathcal{C}$ of size $64 \times 64$ as input and encodes it in the latent feature space. The decoder takes the feature representation and outputs the restored image $f(\mathcal{C},\textbf{w})$. Our encoder and decoder are connected through a channel-wise fully-connected layer. The loss function we use during training is L2 loss, which is the mean squared error between the restored image $f(\mathcal{C},\textbf{w})$ and the real image $\mathcal{R}$. We use a symmetric encoder-decoder pipeline that is efficient for training and effective for learning. It is a unified framework that can be used for all four image restoration tasks. Please see Supp.~for the complete network architecture and detailed design choices.
%===========================================================
%%%%%%%%% EXPERIMENTS
\vspace*{-0.05in}

\section{Experiments}\label{sec:expts}
We compare with traditional ``fixated" learners, hard negative mining, multi-task and curriculum methods, and several existing methods in the literature~\cite{pathak2016context,aharon2006img,dabov2007image,burger2012image,gu2014weighted,schmidt2014shrinkage,chen2016trainable}.

\vspace*{-0.05in}

\subsection{Datasets}

\vspace*{-0.05in}

We experiment with three datasets: CelebFaces Attributes (CelebA)~\cite{liu2015faceattributes}, SUN397 Scenes~\cite{xiao2014sun}, and the Denoising Benchmark 11 (DB11)~\cite{dabov2007image,burger2012image}.  We do not use any of the accompanying labels. For CelebA, we use the first 100,000 images as the training set. Among the rest of the images, we hold out 1,000 images each for the validation and test sets. For SUN397, similarly, we use 100,000 images for training, and 1,000 each for validation and testing. DB11 consists of 11 standard benchmark images, such as ``Lena" and ``Barbara", that have been widely used to evaluate denoising algorithms~\cite{dabov2007image,burger2012image}. We only use this dataset to facilitate comparison with prior work.

\vspace*{-0.05in}
\subsection{Implementation Details}\label{sec:details}
\vspace*{-0.05in}

Our image restoration pipeline is implemented in Torch\footnote{\url{https://github.com/rhgao/on-demand-learning}}. We use ADAM~\cite{kingma2015adam} as the stochastic gradient descent solver. We use the default solver hyper-parameters suggested in~\cite{radford2016unsupervised} and batch size $\mathbb{B}=100$ in all experiments. 

The number of sub-tasks $N$ for on-demand learning controls a trade-off between precision and run-time. Larger values of $N$ will allow the on-demand learning algorithm more fine-grained control on its sample generation, which could lead to better results. However, the time complexity for validating on all sub-tasks at the end of each epoch is $O(N)$. Therefore, a more fine-grained division of training examples among sub-tasks comes at the cost of longer running time during training. For consistency, we divide each of the image restoration tasks into $N=5$ difficulty levels during training. We have not tried any other values, and it is possible other settings could improve our results further. We leave how to select the optimal value of $N$ as future work. An extra level (level 6) is added during testing. The level 6 sub-task can be regarded as an ``extra credit" task that strains the generalization ability of the obtained model.

\vspace*{-0.22in}

\label{exp:subtasks}
\paragraph{\textbf{Image Inpainting:}}
We focus on inpainting missing square blocks of size $1\times1$ to $30\times30$ at different locations across the image. We divide the range into the following five intervals, which define the five difficulty levels: $1 \times 1 - 6\times 6$, $7 \times 7 - 12\times 12$, $13 \times 13 - 18\times 18$, $19 \times 19 - 24\times 24$, $25 \times 25 - 30 \times 30$.

\vspace*{-0.22in}

\paragraph{\textbf{Pixel Interpolation:}}
We train the pixel interpolation network with images corrupted by removing a random percentage of pixels. The percentage is sampled from the range $[0\%,75\%]$. We divide the range into the following five difficulty levels: $0\% - 15\%$, $15\% - 30\%$, $30\% - 45\%$, $45\% - 60\%$, $60\% - 75\%$.

\vspace*{-0.22in}

\paragraph{\textbf{Image Deblurring:}}
Blur kernel widths $\sigma_x$ and $\sigma_y$, which are sampled from the range $[0,5]$, control the level of difficulty.
We consider the following five difficulty levels: $0 - 1$, $1 - 2$, $2 - 3$, $3 - 4$, $4 - 5$.

\vspace*{-0.22in}
\paragraph{\textbf{Image Denoising:}}
We use gray-scale images for denoising. The variance $\sigma$ of additive white Gaussian noise is sampled from the range [0,100]. We use the following five difficulty levels: $0 - 20$, $20 - 40$, $40 - 60$, $60 - 80$, $80 - 100$. 

\vspace*{-0.05in}

\subsection{Baselines}

\vspace*{-0.05in}

For fair comparisons, all baseline models and our method are trained for the same amount of time (1500 epochs). Therefore, while our algorithm shifts the distribution of training instances it demands on the fly, it \emph{never receives more training instances than the baselines}.

\begin{table*}
\centering
\fontsize{6.3}{8}\selectfont
\begin{tabular}{c?{0.5mm}c|l|c|c|l|c|c|c?{0.5mm}c|c|c|c|c|c}
\specialrule{.15em}{.1em}{.1em}
\multirow{3}{*}{}          & \multicolumn{8}{c?{0.5mm}}{CelebA}                                                                                                          & \multicolumn{6}{c}{SUN397}                                                                                              \\ \cline{2-15}
                           & \multicolumn{3}{c|}{Image Deblurring}        & \multicolumn{3}{c|}{Pixel Interpolation}      & \multicolumn{2}{c?{0.5mm}}{Image Inpainting} & \multicolumn{2}{c|}{Image Deblurring} & \multicolumn{2}{c|}{Pixel Interpolation} & \multicolumn{2}{c}{Image Inpainting} \\ \cline{2-15}
                           & \multicolumn{2}{c|}{L2 Loss} & PSNR          & \multicolumn{2}{c|}{L2 Loss} & PSNR           & L2 Loss             & PSNR            & L2 Loss             & PSNR            & L2 loss              & PSNR              & L2 Loss             & PSNR            \\ \hline
Rigid Joint Learning       & \multicolumn{2}{c|}{1.58 }        &    29.40 dB         & \multicolumn{2}{c|}{1.02 }        &    31.86 dB            &       1.05              &  32.11 dB              &    2.32                  &      28.53 dB           &   1.29                    &      31.98 dB             &   1.80                   &     31.13 dB            \\ \hline
Cumulative Curriculum      & \multicolumn{2}{c|}{1.85 }        &    28.70 dB           & \multicolumn{2}{c|}{1.11 }        &    31.68 dB            &       1.28               &  31.47 dB                &       2.64               &  27.86 dB               &      1.36                 &    31.70 dB               &      1.94                &   30.75 dB              \\ \hline
Cumulative Anti-Curriculum & \multicolumn{2}{c|}{1.49 }        &    29.31 dB           & \multicolumn{2}{c|}{1.01 }        &    31.96 dB            &    1.04              &     31.90 dB            &      2.39                &    28.34 dB              &     1.25                &      32.02 dB             &     1.90                 &    30.44 dB             \\ \hline
Staged Curriculum          & \multicolumn{2}{c|}{125 }        &    15.59 dB           & \multicolumn{2}{c|}{2.10 }        &    28.51 dB            &       1.18          &   31.30 dB           &      133                &    14.44 dB             &     2.36                  &       28.13 dB            &                     1.87  &     30.42 dB            \\ \hline
Staged Anti-Curriculum     & \multicolumn{2}{c|}{5.54 }        &    25.43 dB           & \multicolumn{2}{c|}{7.76 }        &    27.82 dB            &         4.80            &   28.10 dB              &      6.27                &  25.17 dB               &        7.05               &    27.76 dB               &      4.35               &   28.42 dB              \\ \hline
Hard Mining     & \multicolumn{2}{c|}{2.98 }        &    27.33 dB           & \multicolumn{2}{c|}{1.85 }        &    29.15 dB            &         3.31            &   29.47 dB              &      3.98                &  26.35 dB               &        1.82               &    29.01 dB               &      2.61              &   29.83 dB              \\ \hline
On-Demand Learning  & \multicolumn{2}{c|}{\textbf{1.41}}        &  \textbf{29.58 dB}             & \multicolumn{2}{c|}{\textbf{0.95} }        &    \textbf{32.09 dB}            &       \textbf{0.99}               &    \textbf{32.30 dB}            &    \textbf{2.11}                   &     \textbf{28.70 dB}            &  \textbf{1.19}                     &        \textbf{32.21 dB}           &   \textbf{1.69}                   &      \textbf{31.38 dB}           \\
\specialrule{.15em}{.1em}{.1em}

\end{tabular}
\caption{Summary of the overall performance of all algorithms for three image restoration tasks on the CelebA and SUN397 datasets. (See Supp.~for similar results on denoising). Overall performance is measured by the mean L2 loss (in \textperthousand, lower is better) and mean PSNR (higher is better) averaged over all sub-tasks.  \KG{Numbers are obtained over 20 trials with standard error (SE) approximately $5\times10^{-6}$ for L2 loss and $3\times10^{-3}$ for PSNR on average.} \RG{A paired t-test shows the results are significant with p-value $5\times10^{-30}$.}}

\label{Table:overall}
\vspace*{-0.15in}

\end{table*}	
\begin{figure}
  \centering
  \includegraphics[scale=0.53]{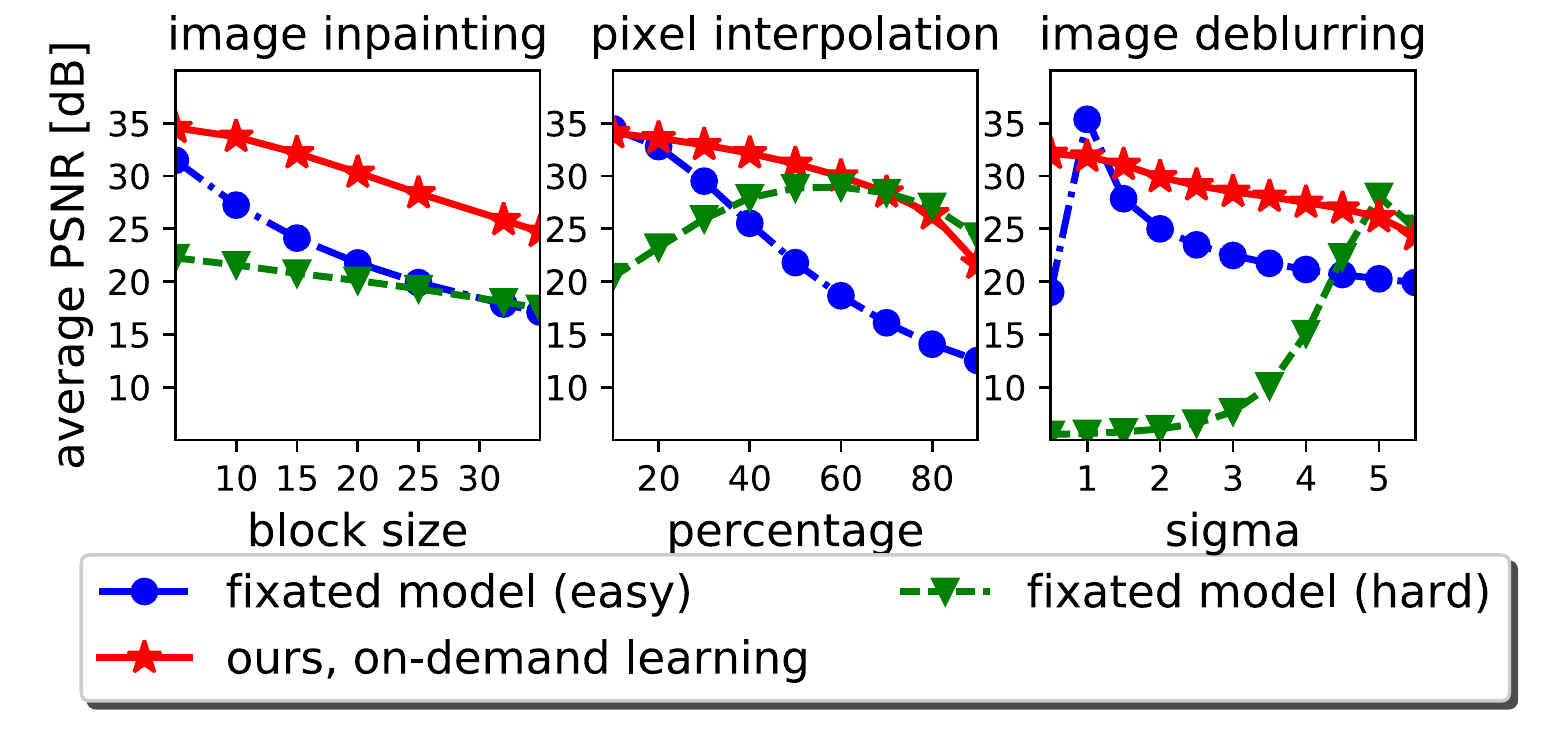}
  \caption{Our algorithm vs.~fixated models on CelebA (See Supp.~for results on SUN397 and denoising). Our algorithm performs well over the spectrum of difficulty, whereas fixated models perform well at only a certain level of corruption.}
  \label{fig:compare2pony}
   \vspace*{-0.2in}
\end{figure}

\vspace*{-0.15in}

\label{baselines}
\paragraph{\textbf{Fixated Model (Hard):}}
The image restoration network is trained only on one level of severely corrupted images.

\vspace{-0.2in}

\paragraph{\textbf{Fixated Model (Easy):}}
The image restoration network is trained only on one level of lightly corrupted images.

\vspace{-0.2in}

\paragraph{\textbf{Rigid Joint Learning:}}
The image restoration network is trained on all sub-tasks of different difficulty levels (level 1-$N$) jointly. We allocate the same number of training examples for each sub-task per batch.

\vspace{-0.2in}

\paragraph{\textbf{Staged Curriculum Learning:}}
The network starts at the easiest sub-task (level 1) and gradually switches to more difficult sub-tasks. At any time, the network trains on only one sub-task. It trains on each sub-task for 300 epochs.

\vspace{-0.2in}

\paragraph{\textbf{Staged Anti-Curriculum Learning:}}
The network performs as the above, but reverses the curriculum to start with the most difficult task (level $N$).

\vspace{-0.2in}

\paragraph{\textbf{Cumulative Curriculum Learning:}}
The network starts at the easiest sub-task (level 1) and gradually adds more difficult sub-tasks and learns them jointly. More specifically, the baseline model is first trained on level 1 sub-task for 300 epochs, and then performs rigid joint learning on sub-tasks of level 1 and 2 for 300 epochs, \cc{followed by performing rigid joint learning on sub-tasks of level 1,2,3 for another 300 epochs,} and so on.

\vspace{-0.2in}

\paragraph{\textbf{Cumulative Anti-Curriculum Learning:}}
The network performs as the above, but reverses the curriculum.
\vspace{-0.2in}

\paragraph{\textbf{Hard Mining:}} For each task, we create a dataset of 1M images with various corruptions. We directly train on the dataset for 50 epochs, then continue training with hard mining until convergence. To select hard examples, we identify those with the largest reconstruction loss and use them to compute and back propagate gradients. Specifically, in each batch, we select the 10 with highest loss. 

\KG{As far as source training data,} the fixated model baselines represent the status quo in using deep learning for image restoration tasks~\cite{liu2016learning,pathak2016context,yeh2016semantic,xie2012image,jain2009natural,burger2012image,schuler2013machine}, while the rigid joint learning baseline represents the natural solution of pooling all training data~\cite{jain2009natural,mao2016image}.  The curriculum methods are of our own design. The hard mining baseline is designed to best mimic traditional hard negative mining strategies.
\KG{Our system never receives more training images than any baseline; only the distribution of distortions among those images evolves over epochs.} We test all algorithms across the whole spectrum of difficulty (sub-task 1-$N$ and an extra level), and synthesize corresponding testing instances randomly over 20 trials. No methods have prior knowledge of the test distribution, thus none are able to benefit from better representing the expected test distribution during training.

\subsection{\KG{Fixated Model vs.~Our Model}}
\RG{We first show that our on-demand algorithm successfully addresses the fixation problem, where the fixated models employ an identical network architecture to ours.}  
For inpainting, the fixated model (hard/easy) is only trained to inpaint $32 \times 32$ or $5 \times 5$ central blocks, respectively; for pixel interpolation, 80\% (hard) or 10\% (easy) pixels are deleted; for deblurring, $\sigma_x=\sigma_y=5$ (hard) or $\sigma_x=\sigma_y=1$ (easy);
for denoising,  $\sigma=90$ (hard) or $\sigma=10$ (easy).

\begin{figure*}
  \centering
  \includegraphics[scale=0.82]{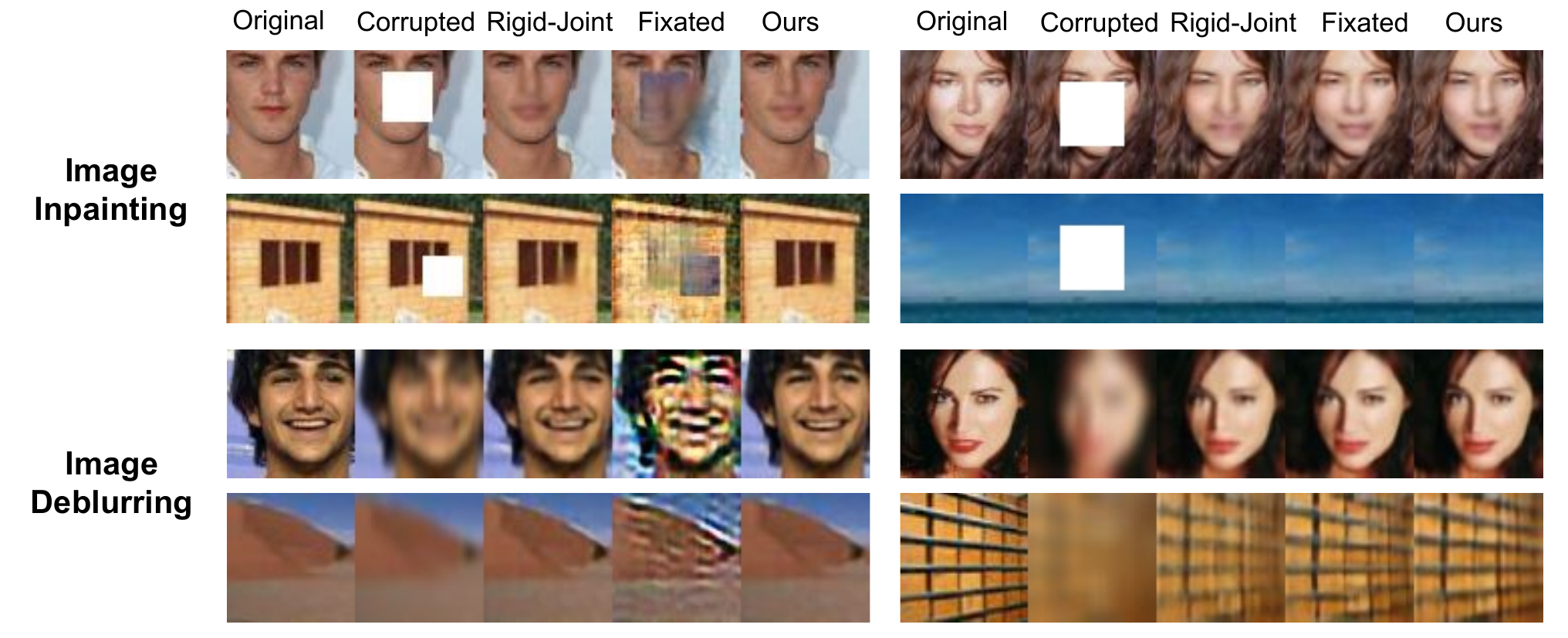}
  \caption{For each task, the first row shows testing examples of CelebA dataset, and the second row shows examples of SUN397 dataset. While the fixated model can only perform well at one level of difficulty (right col), the all-rounder models trained using our proposed algorithm perform well on images with various corruption levels. See Supp.~for similar results on pixel interpolation and image denoising.}
\label{fig:qualitative}

\vspace*{-0.1in}

\end{figure*}

Fig.~\ref{fig:compare2pony} summarizes the test results on images of various corruption levels on CelebA (See Supp.~for all). The fixated model overfits to a specific corruption level (easy or hard). It succeeds beautifully for images within its specialty (e.g., \RG{the sudden spike in Fig.~\ref{fig:compare2pony}} (right)), but performs poorly when forced to attempt instances outside its specialty. For inpainting, the fixated models also overfit to the central location, and thus cannot perform well over the whole spectrum. In contrast, models trained using our algorithm perform well across the spectrum of difficulty.

\subsection{Comparison to Existing Inpainter}
We also compare our image inpainter against a state-of-the-art inpainter from Pathak~\etal~\cite{pathak2016context}. We adapt their provided code\footnote{\url{https://github.com/pathak22/context-encoder}} and follow the same procedures as in~\cite{pathak2016context} to train two variants on CelebA: one is only trained to inpaint central square blocks, and the other is trained to inpaint regions of arbitrary shapes using random region dropout. Table~\ref{exp:cmp_berkeley} compares both variants to our model on the held out CelebA test set. Their first inpainter performs very well when testing on central square blocks (left cols), but it is unable to produce satisfactory results when tested on square blocks located anywhere in the image (right cols). Their second model uses random region dropout during training, but our inpainter still performs much better. The ``all-rounder" inpainter trained under our on-demand learning framework does similarly well in both cases.  It is competitive---and stronger on the more difficult task---even without the use of adversarial loss as used in their framework during training. Please also see Supp.~for some real-world applications (e.g., object removal in photos).

\begin{table}[ht!]
\vspace*{-0.1in}
\fontsize{6.5}{8}\selectfont
\centering
\begin{tabular}{c?{0.3mm}cc?{0.3mm}cc}
\specialrule{.12em}{.1em}{.1em}
\multirow{2}{*}{Method} & \multicolumn{2}{c?{0.3mm}}{Central Square Block} & \multicolumn{2}{c}{Arbitrary Square Block} \\ \cline{2-5}
                   & L2 Loss       & PSNR             & L2 Loss        & PSNR       \\ \hline
Pathak~\etal~\cite{pathak2016context} Center         &  \textbf{0.83\%}\            &                \textbf{22.16 dB}                  &    6.84\%           &     11.80 dB       \\
Pathak~\etal~\cite{pathak2016context} +Rand drop           &   2.47\%            &         16.18  dB             &    2.51\%            &     16.20 dB       \\
Ours                      &  0.93\%                        &   20.74 dB        &   \textbf{1.04\%}          &    \textbf{20.31 dB}                        \\ \hline
\end{tabular}
\caption{Image inpainting accuracy for CelebA on two test sets.}
\vspace*{-0.2in}
\label{exp:cmp_berkeley}
\end{table}
\subsection{\RG{On-Demand Learning vs.~Alternative Models}}

We next compare our method to the hard mining, curriculum and multi-task baselines. Table~\ref{Table:overall} shows the results (Please see Supp.~for similar results on image denoising). We report average L2 loss and PSNR over all test images. Our proposed algorithm consistently outperforms the \cc{well-designed} baselines. \RG{Hard mining overfits to the hard examples in the static pool of images, and the Staged (Anti-)Curriculum Learning algorithms overfit to the last sub-task they are trained on, yielding inferior overall performance. The Cumulative (Anti-)Curriculum Learning algorithms and Rigid Joint Learning are more competitive, because they learn sub-tasks jointly and try to perform well on sub-tasks across all difficulty levels. However, the higher noise levels dominate their training procedure by providing stronger gradients. As training goes on, these methods cannot provide the optimal distribution of gradients across corruption levels for effective learning.} By automatically guiding the balance among sub-tasks, our algorithm obtains the best all-around performance. Especially, we observe our approach generalizes better to difficulty levels never seen before, and performs better on the ``extra credit" sub-task.

Fig.~\ref{fig:qualitative} shows qualitative examples output by our method for inpainting and deblurring. See Supp.~for similar results of interpolation and denoising. These illustrate that models trained using our proposed on-demand approach perform well on images of different degrees of corruption. \emph{With a single model}, we inpaint blocks of different sizes at arbitrary locations, restore corrupted images with different percentage of deleted pixels, deblur images at various degrees of blurriness, and denoise images of various noise levels. In contrast, the fixated models can only perform well at one level of difficulty that they specialize in. Even though we experiment with images of small scale ($64 \times 64$) for efficiency, qualitative results of our method are still visually superior to other baselines including rigid-joint learning.

\label{exp:extra_data}
We argue that the gain of our algorithm does not rest on \emph{more} training instances of certain sub-tasks, but rather a suitable combination of sub-tasks for effective training. Indeed, we never use more training instances than any baseline. To emphasize this point, we separately train a rigid-joint learning model using 200,000 training images (the original 100,000 and the \emph{extra} 100,000) from CelebA. \footnote{The other datasets lack sufficient data to run this test.}
We observe that the extra training instances do not help rigid joint training converge to a better local minimum. This result suggests on-demand learning's gains persist even if our method is put at the disadvantage of having access to 50\% fewer training images.

How does the system focus its attention as it learns?  To get a sense, we examine the learned allocation of sub-tasks during training.  
Initially, each sub-task is assigned the same number of training instances per batch.  \KG{In all tasks, as training continues, the network tends to dynamically shift its allocations to put more emphasis on the ``harder" sub-tasks, while never abandoning the ``easiest" ones.} \RG{The right proportions of difficulty lead to the superior overall performance of our model.}

\vspace*{-0.05in}

\subsection{Comparison to Existing Denoising Methods}

\vspace*{-0.05in}

\label{exp:denoising}

\begin{table}
\centering
\fontsize{6}{7.5} \selectfont
\begin{tabular}{c?{0.5mm}c|c|c|c|c|c?{0.3mm}c}
Image   & \cite{aharon2006img}  & \cite{dabov2007image} & \cite{burger2012image}    & \cite{gu2014weighted} & \cite{schmidt2014shrinkage} & \cite{chen2016trainable} & Ours \\ \specialrule{.12em}{.1em}{.1em}
Barbara & 29.49 & 30.67 & 29.21 & \textbf{31.24} & 28.95 &  29.41 & 28.92 / 29.63 \\ \hline
Boat    & 29.24 & 29.86 & 29.89 & 30.03 & 29.74 & 29.92  & \textbf{30.11} / \textbf{30.15}   \\ \hline
C.man   & 28.64 & 29.40 & 29.32  & 29.63 & 29.29 & 29.71 & 29.41 / \textbf{29.78}  \\ \hline
Couple  & 28.87 & 29.68 & 29.70 & 29.82 & 29.42 & 29.71  & \textbf{30.04} / \textbf{30.02}   \\ \hline
F.print & 27.24 & 27.72 & 27.50 & \textbf{27.88} & 27.02 & 27.32  & 27.81 / 27.77  \\ \hline
Hill   & 29.20 & 29.81 & 29.82 & 29.95 & 29.61 & 29.80  & \textbf{30.03} / \textbf{30.04}   \\ \hline
House   & 32.08 & 32.92 & 32.50 & \textbf{33.22} & 32.16 & 32.54 & 33.14 / 33.03  \\ \hline
Lena   & 31.30 & 32.04 & 32.12 & 32.24 & 31.64 & 32.01 & \textbf{32.44} / \textbf{32.36}  \\ \hline
Man    & 29.08 & 29.58 & 29.81 & 29.76 & 29.67 & 29.88 & \textbf{29.92} / \textbf{29.96}   \\ \hline
Montage  & 30.91 & 32.24 & 31.85 & 32.73 & 31.07 & 32.29 & 32.34 / \textbf{32.74}  \\ \hline
Peppers & 29.69 & 30.18 & 30.25 & 30.40 & 30.12 & \textbf{30.55} & 30.29 / 30.48  \\
\end{tabular}
\caption{PSNRs (in dB, higher is better) on standard test images, $\sigma = 25$. We show the performance of both our all-rounder model (left) and fixated model (right) of our image denoising system. Note that our on-demand learning model is the \emph{only} one that \emph{does not} exploit the noise level ($\sigma$) of test images.}

\vspace*{-0.2in}
\label{exp:denoising_table}
\end{table}

In previous sections, we have compared our on-demand learning denoising model with alternative models. To facilitate comparison to prior work and demonstrate the competitiveness of our image restoration framework, in this section we perform a case study on the image denoising task using our denoising system. See Supp.~for details about how we denoise images of arbitrary sizes.

We test our image denoising system on DB11~\cite{dabov2007image,burger2012image}. We first compare our model with state-of-the-art denoising algorithms on images with a specific degree of corruption ($\sigma = 25$, commonly adopted to train fixated models in the literature). Table~\ref{exp:denoising_table} summarizes the results\footnote{We take the reported numbers \cite{burger2012image} or use the authors' public available code~\cite{gu2014weighted,schmidt2014shrinkage,chen2016trainable} to generate the results in Table~\ref{exp:denoising_table}.}.  \RG{Although using a simple encoder-decoder network, we still have very competitive performance. Our on-demand learning model outperforms all six existing denoising algorithms on 5 out of the 11 test images (7 out of 11 for the fixated version of our denoising system), and is competitive on the rest.} Note that our on-demand learning model \emph{does not need to know the noise level of test images}. However, all other compared algorithms either have to know the exact noise level ($\sigma$ value), or train a separate model for this specific level of noise ($\sigma = 25$).

\KG{More importantly}, the advantage of our method is more apparent when we test across the spectrum of difficulty levels.  We corrupt the DB11 images with AWG noise of increasing \KG{magnitude} and compare with the denoising algorithms BM3D~\cite{dabov2007image} and MLP~\cite{burger2012image} based on the authors' public code\footnote{\url{http://www.cs.tut.fi/~foi/GCF-BM3D/}}\footnote{\url{http://people.tuebingen.mpg.de/burger/neural_denoising/}} and reported results~\cite{burger2012image}.
 We compare with two MLP models: one is trained only on corrupted images of  $\sigma = 25$, and the other is trained on images with various noise levels. BM3D and MLP both need to be provided with the correct level of the noise ($\sigma$) during testing. We also run a variant of BM3D for different noise levels but fix the specified level of noise to $\sigma = 25$ .

\begin{figure}
  \centering
  \includegraphics[scale=0.30]{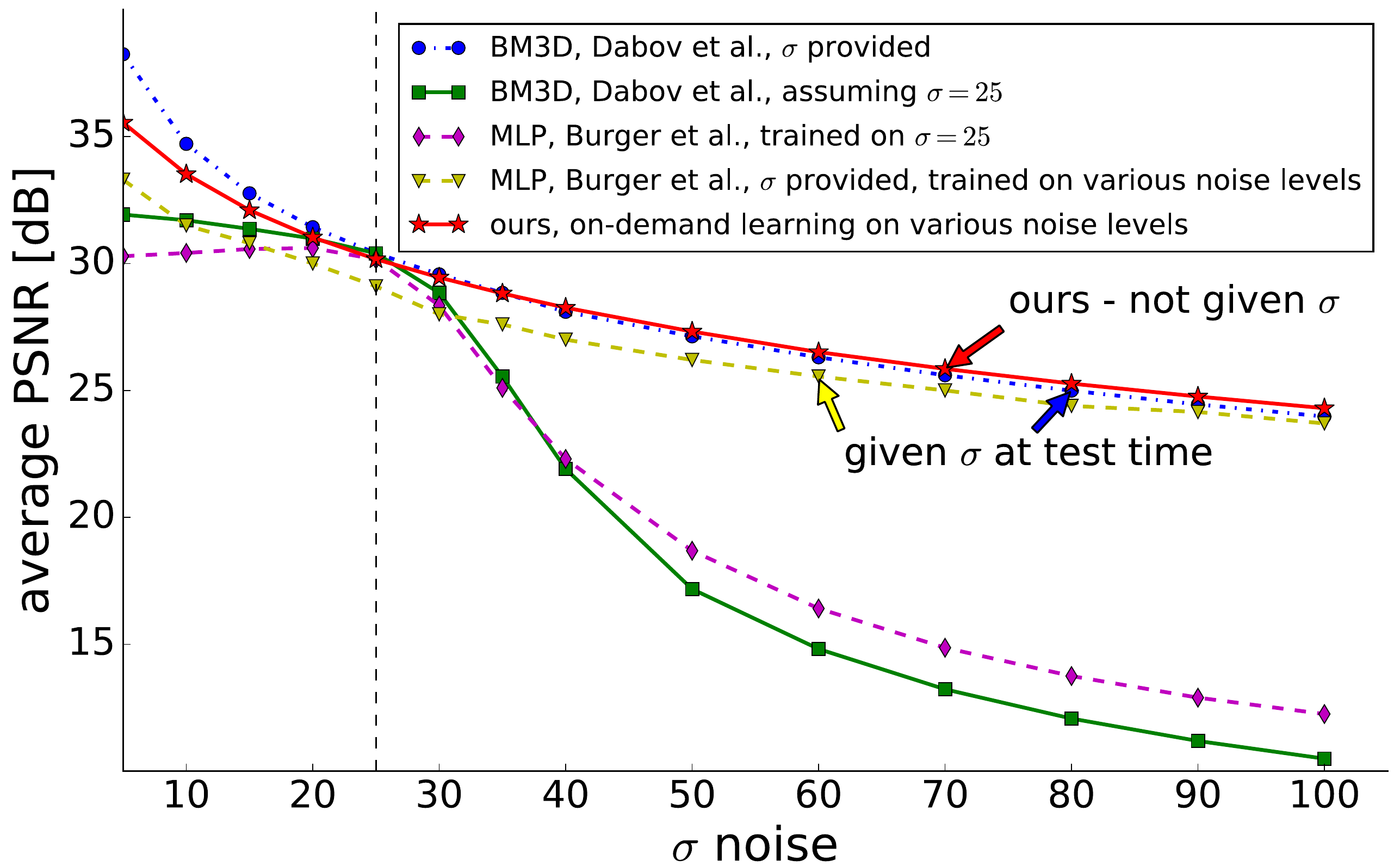}
  \caption{Comparisons of the performance of image denoising systems at different noise levels. Our system is competitive over the whole spectrum of noise levels without \KG{requiring} knowledge of the corruption level of test images.  \KG{Best viewed in color.}}
  \label{fig:denoise_compare}
  \vspace*{-0.2in}
\end{figure}

Fig.~\ref{fig:denoise_compare} shows the results.  We see that the MLP model~\cite{burger2012image} trained on a single noise level only performs well at that specific level of corruption. Similarly, BM3D~\cite{dabov2007image} needs the correct input of noise level in order to perform well across the spectrum of noise levels. In contrast, our image denoising system consistently performs well on all noise levels, yet we \emph{do not} assume knowledge of $\sigma$ during testing.  \KG{This is an essential advantage for real-world applications.}
%===========================================================
%%%%%%%%% CONCLUSION
\vspace*{-0.2in}

\section{Conclusion}

\vspace*{-0.05in}

We have addressed a common problem in existing work that leverages deep models to solve image restoration tasks: overfitting. We devise a symmetric encoder-decoder network amenable to all image restoration tasks, and propose a simple but novel \emph{on-demand learning} algorithm that turns a fixated model into one that performs well on a task across the spectrum of difficulty. Experiments on four tasks on three diverse datasets demonstrate the effectiveness of our method. 
Our on-demand learning idea is a general concept not restricted to image restoration tasks, and may be applicable in other domains as well, e.g., self-supervised feature learning. As future work, we plan to design continuous sub-tasks to avoid discrete sub-task bins, and we will explore ways to make an image restoration task more self-paced by allowing the network to design the most desired sub-task on its own. \RG{Finally, another promising direction is to explore combinations of different types of distortions.}
%===========================================================

\noindent\textbf{Acknowledgements:} This research is supported in part by NSF IIS-1514118. We also gratefully acknowledge the support of the Texas Advanced Computing Center (TACC) and a GPU donation from Facebook.

{\small
\bibliographystyle{ieee}
\bibliography{cv_archive}
}

\clearpage
\newpage
{\LARGE \textbf{Appendix}}
\linebreak
\appendix

\noindent The supplementary materials consist of:
\begin{enumerate}
	\item[A.] Pseudocode for our on-demand learning algorithm.
	\item[B.] Details of our network architecture.
	\item[C.] Details of the fixated models setup.
	\item[D.] Fixated models vs. All-rounder on SUN397 and image denoising.
	\item[E.] Overall performance of our image denoising model.
	\item[F.] Applications of our image inpainter to real images.
	\item[G.] Qualitative results for interpolation and denoising.
	\item[H.] Image denoising qualitative results on DB11.
\end{enumerate}
%===========================================================
\section{On-Demand Learning Algorithm}
We present the pseudocode of our on-demand learning algorithm as follows:
\begin{algorithm}
\label{algorithm}
\caption{On-Demand Learning}
 \textbf{N sub-tasks of increasing difficulty:} $T_1, T_2, \ldots, T_{N}$\\
 \textbf{\# of training examples for sub-task $T_i$ per batch:} $B_i$\\
 \textbf{Batch Size:} $\mathbb{B}$\\
 \textbf{Initialization:} $B_i = \mathbb{B}/N$ \\
 \While{not converge}{
  continue training for one epoch and snapshot;\\
  \If{end of epoch}{
  	 $i = 1$;\\
  	 \For{$i \leq N$}{
  	 validate snapshot model on sub-task $T_i$;\\
  	 get mean PSNR $P_i$;}
  	 update $B_i = \frac{1/P_i}{\sum_{i=1}^{N}1/P_i} \cdot \mathbb{B}$;
   }{
  }
 }
\end{algorithm}

%===========================================================
\section{Deep Learning Network Architecture}
Fig.~\ref{fig:network} shows the complete network architecture used for all tasks to implement our on-demand learning idea. Our image restoration network is a symmetric encoder-decoder pipeline.  The encoder takes a corrupted image of size $64 \times 64$ as input and encodes it in the latent feature space. The decoder takes the feature representation and outputs the restored image. Our encoder and decoder are connected through a channel-wise fully-connected layer.

Specifically, for our encoder, we use four convolutional layers. Following similar design choices in DCGAN~\cite{radford2016unsupervised}, we put a batch normalization layer~\cite{ioffe2015batch} after each convolutional layer to accelerate training and stabilize learning. The leaky rectified linear unit (LeakyReLU) activation~\cite{maas2013rectifier,xu2015empirical} is used in all layers in the encoder. 

The four convolutional layers in the encoder only connect all the feature maps together, but there are no direct connections among different locations within each specific feature map. Fully-connected layers are usually used to handle this information propagation in present successful network architectures~\cite{krizhevsky2012imagenet,simonyan2014very}. In our network, the latent feature dimension is $4 \times 4 \times 512 = 8192$ for both encoder and decoder. Fully-connecting our encoder and decoder will increase the number of parameters explosively. To more efficiently train our network and demonstrate our concept, we use a channel-wise fully-connected layer to connect the encoder and decoder, as in~\cite{pathak2016context}. The channel-wise fully-connected layer is designed to only propagate information within activations of each feature map. In our case, each $4 \times 4$ feature map in the encoder side is fully-connected with each $4 \times 4$ feature map in the decoder side. This largely reduces the number of parameters in our network and accelerates training significantly. 

The decoder consists of four up-convolutional layers~\cite{long2015fully,dosovitskiy2015learning,zeiler2014visualizing}, each of which is followed by a rectified linear unit (ReLU) activation except the output layer. We use the Tanh function in the output layer, and the output is of the same size as the input image. The series of up-convolutions and non-linearities conducts a non-linear weighted upsampling of the feature produced by the encoder and generates a higher resolution image of our target size ($64 \times 64$).

\begin{figure*}
  \centering
  \includegraphics[scale=1.2]{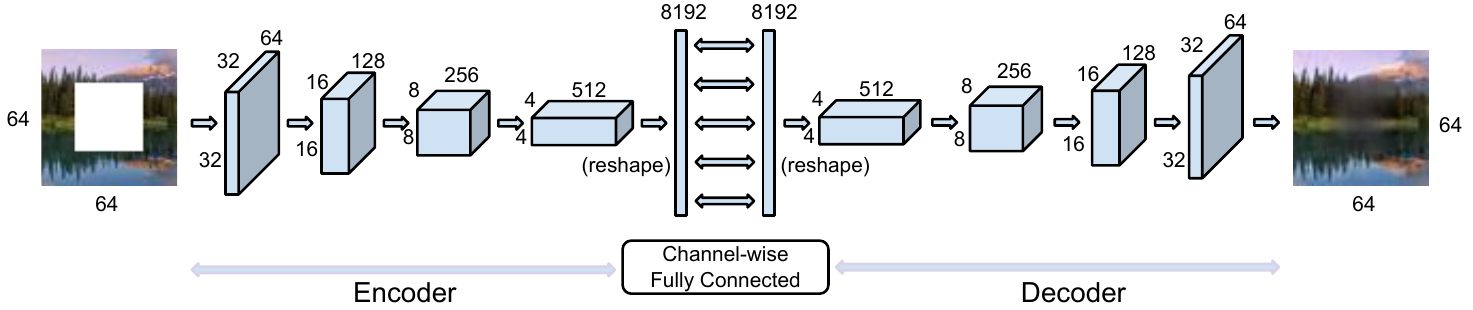}
  \caption{Network architecture for our image restoration framework. Our image restoration framework is an encoder-decoder pipeline with the encoder and decoder connected by a channel-wise fully-connected layer. The illustration is for image inpainting task. The same network architecture also holds for the other three tasks: pixel interpolation, image deblurring, and image denoising.}
  \label{fig:network}
\end{figure*}

%===========================================================
\section{Details of the Fixated Models Setup}

\begin{figure}
  \centering
  \includegraphics[scale=0.48]{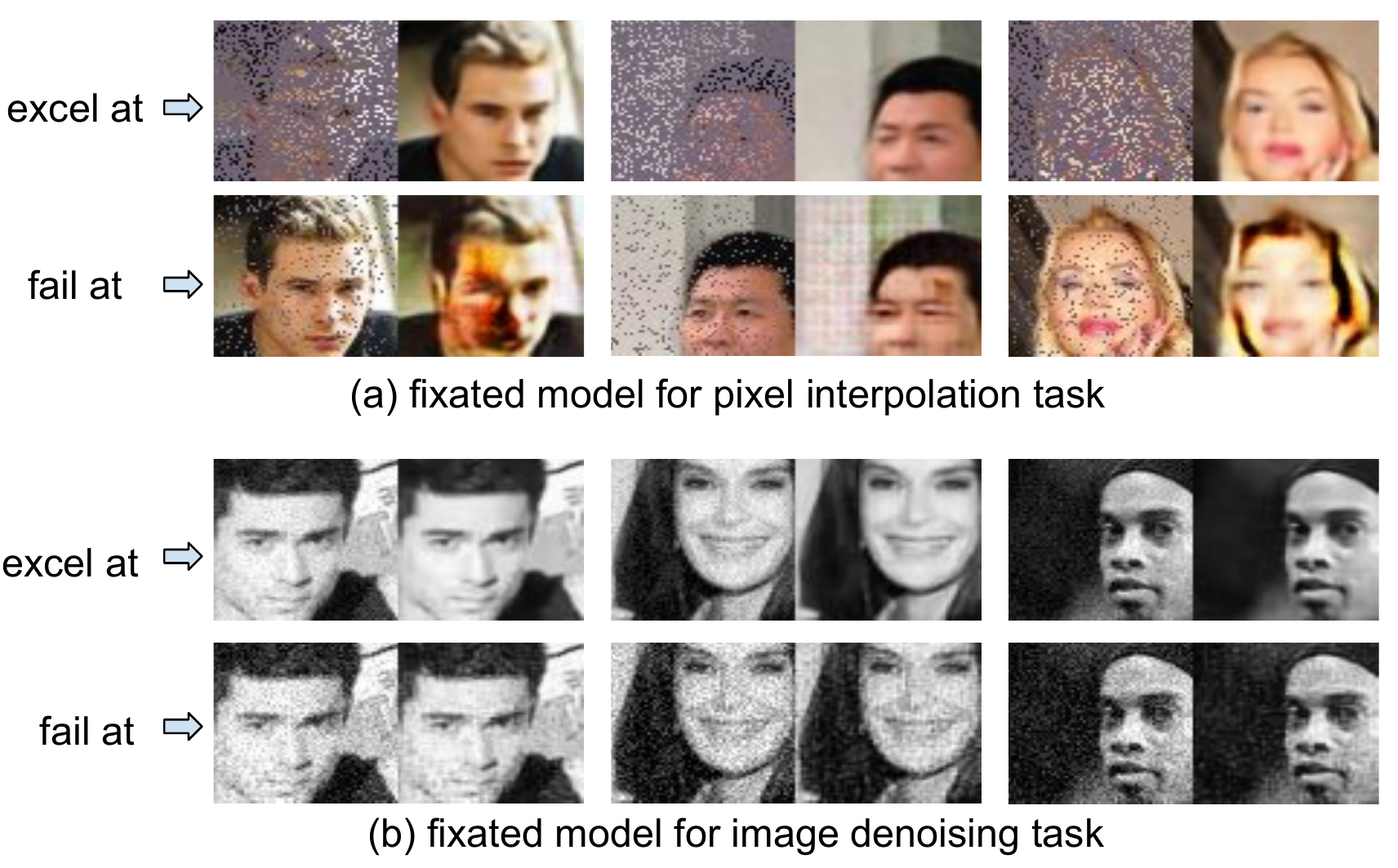}
  \caption{Qualitative examples of fixated models for pixel interpolation and image denoising tasks. The models overfit to a certain degree of corruption. They perform extremely well at that level of corruption, yet fail to produce satisfactory restoration results even for much easier sub-tasks.}
  \label{fig:overfitting}
\end{figure}

We followed the current literature to train deep networks to target a certain degree of corruption for four tasks---image inpainting, pixel interpolation, image deblurring and image denoising---and demonstrate how severe the fixation problem is. We show the qualitative examples of fixated models for pixel interpolation and image denoising tasks in Fig.~\ref{fig:overfitting} as a supplement to Fig.~2 in the main paper.

Specifically, \textbf{for the image inpainting task}, we follow similar settings in~\cite{pathak2016context,yeh2016semantic} and train a model to inpaint a large central missing block of size $32 \times 32$. During testing, the resulting model can inpaint the central block of the same size at the same location very well (first row in Fig.~2-a in the main paper). However, if we remove a block that is slightly shifted away from the central region, or remove a much \emph{smaller} block, the model fails to inpaint satisfactorily (second row in Fig.~2-a in the main paper). Following~\cite{pathak2016context}, we replace pixels in removed blocks with the average pixel values in training images (which tend to look grey). We can observe that grey areas are retained in regions outside of the central block in the failure cases, which is a strong indicator that the trained network severely overfits to the central location.

\textbf{For the pixel interpolation task}, we train a model only based on heavily corrupted images (80\% of random pixels deleted), following~\cite{yeh2016semantic}. During testing, if we use the obtained model to restore images of the same corruption level, the images are recovered very well (first row in Fig.~\ref{fig:overfitting}-a). However, if we test the same model on \emph{lightly} corrupted (easier) images, the model performs very poorly (second row in Fig.~\ref{fig:overfitting}-a). The trained network either produces common artifacts of deep networks like the checkerboard artifacts, or a much blurrier low-quality restored image.

\textbf{For the image deblurring task}, results are similar. We train a model only based on heavily blurred images ($\sigma_x = \sigma_y = 5$). The trained model can successfully restore very blurry images (same blurry level as training examples), but is unable to restore images that are much less blurry. In the second row of Fig.2-b in the main paper, we can observe some ripple artifacts, which may be similar to the shape of the Gaussian kernel function that the network overfits to.

\textbf{For the image denoising task}, we train a model only based on lightly corrupted images ($\sigma = 10$ for AWG noise). During testing, the model can successfully restore images of the same level of noise (first row in Fig.~\ref{fig:overfitting}-b). However, it fails catastrophically when we increase the severity of noise on test images (second row in Fig.~\ref{fig:overfitting}-b).

\begin{figure}
  \centering
  \includegraphics[scale=1.65]{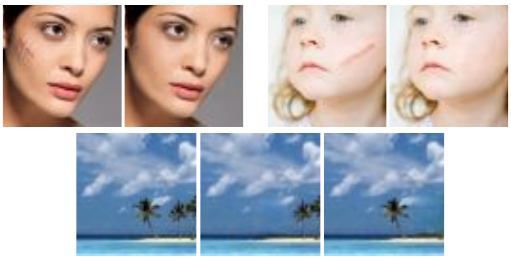}
  \caption{Real applications of our image inpainter. For each example, the left image is the target real image, and the right images are images processed by our image inpainter. Our inpainter can successfully remove scars on human faces, and selectively remove objects (trees in the last example) in photographs. 
}
  \label{fig:real_example}
\end{figure}

%===========================================================

\section{Fixated Models vs.~All-Rounder on SUN397 and Image Denoising}

We show the complete comparison of our algorithm with fixated models on CelebA and SUN397 for all of the four tasks in Fig.~\ref{fig:compare2pony}, as a supplement to Fig. 4 in the main paper, where due to space constraints we could show only the CelebA results for three tasks. Results on SUN397 and image denoising are similar. Fixated models overfit to a specific corruption level (easy or hard). It succeeds beautifully for images within its specialty, but performs poorly when forced to attempt instances outside its specialty. In contrast, models trained using our algorithm perform well across the whole spectrum of difficulty. For inpainting, the fixated models even perform poorly at the size they specialize in, because they also overfit to the central location, thus cannot inpaint satisfactorily at random locations at test time.

\begin{figure*}
  \vspace{0.2in}
  \centering
  \includegraphics[scale=0.28]{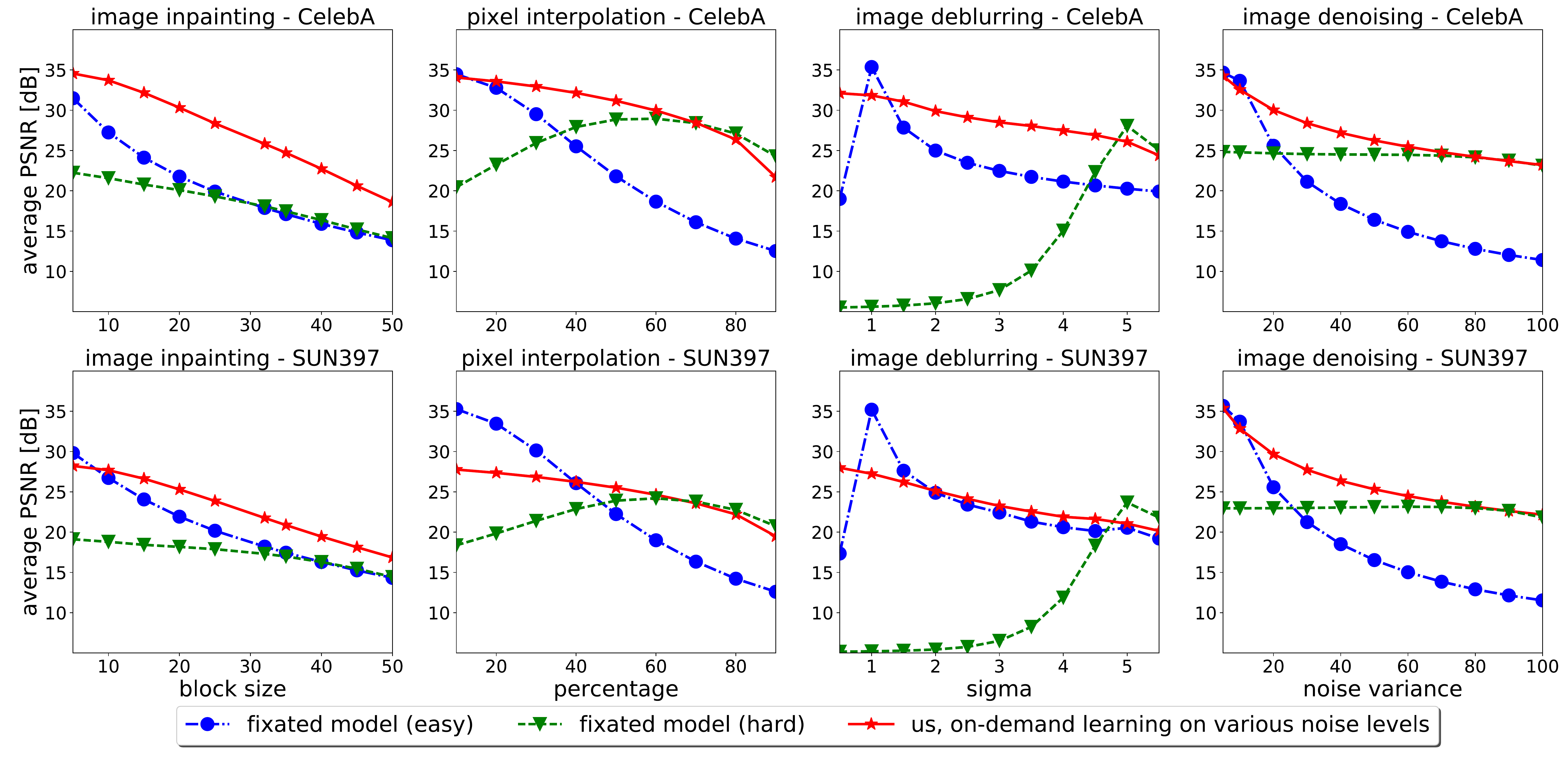}
  \caption{Our on-demand learning algorithm vs.~fixated models for all the four tasks on CelebA and SUN397. This figure is a supplement to Fig.~4 in the main paper, where due to space constraints we could show only the results for three task on CelebA. Models trained using our algorithm perform well over the spectrum of difficulty, while fixated models perform well at only a certain level of corruption.}
  \label{fig:compare2pony}
  \vspace{0.4in}
\end{figure*}
%===========================================================
\section{Overall Performance of Our Image Denoising Model}
In Table~\ref{denoising_table}, we report average L2 loss and PSNR over all test images for the image denoising task, as a supplement to Table~1 in the main paper, where due to space constraints we could show only the results for three tasks. The results for image denoising are similar. Our proposed algorithm consistently outperforms all the well-designed baselines.  

\begin{table}
\centering
\fontsize{7}{8.5}\selectfont
\begin{tabular}{c?{0.5mm}c|c|c|c}
\specialrule{.12em}{.1em}{.1em}
\multicolumn{1}{c?{0.5mm}}{\multirow{2}{*}{}} & \multicolumn{2}{c|}{CelebA Dataset} & \multicolumn{2}{c}{SUN397 Dataset} \\ \cline{2-5} 
\multicolumn{1}{c?{0.5mm}}{}                  & L2 Loss           & PSNR            & L2 Loss           & PSNR            \\ \hline
Rigid Joint Learning                    & 5.90              & 26.38 dB           & 7.56              & 25.69 dB          \\ \hline
Cumulative Curriculum                   &       6.10            &  26.31 dB               & 7.73              & 25.60  dB         \\ \hline
Cumulative Anti-Curriculum              &      5.90             &   26.35 dB             & 7.57              & 25.66  dB         \\ \hline
Staged Curriculum                       & 7.10              & 24.74 dB         & 9.11              & 23.87  dB         \\ \hline
Staged Anti-Curriculum                  &       53.1            &   21.19 dB             &             55.5      &        19.66 dB         \\ \hline
Hard Mining                  &       6.53            &  25.10 dB             &             8.52      &        24.10 dB         \\ \hline
On-Demand Learning                      & \textbf{5.79}              &  \textbf{26.48 dB}          &  \textbf{7.49}              & \textbf{25.80 dB}          \\ \specialrule{.12em}{.1em}{.1em}
\end{tabular}

\vspace*{0.05in}
\caption{Summary of the overall performance of all algorithms for image denoising on CelebA and SUN397. This table is a supplement to Table 1 in the main paper, where due to space constraints we could show only the results for three tasks. Overall performance is measured by the mean L2 loss (in \textperthousand, lower is better) and mean PSNR (higher is better) averaged over all sub-tasks.}
\label{denoising_table}
\end{table}
%===========================================================
\section{Applications of Our Image Inpainter}
We show some applications of our image inpainter to real world scenarios in this section. Fig.~\ref{fig:real_example} shows some examples of using our image inpainter to do scar removal on human face images, and object removal on natural scene images. For each example, the left image is the target real world image. Our inpainter can successfully remove scars on human faces, and selectively remove objects in photographs. 

%===========================================================

\section{Qualitative results for interpolation and denoising}
We show the qualitative examples output by our method for pixel interpolation and image denoising tasks in Fig.~\ref{fig:qualitative} as a supplement to Fig.~5 in the main paper. For each task, the first and second rows show test examples from CelebA and SUN397, respectively. For each quintuple, the first column shows the ground-truth image from the original dataset; the second column shows the corrupted image; the third column shows the restored image using the model trained using rigid joint learning; the fourth column shows the restored image using a fixated model; the last column shows the restored image using the all-rounder model trained by our algorithm. The fixated models can only perform well at a particular level of corruption. Models trained using our proposed on-demand approach are all-rounders that perform well on images of different degrees of corruption. With a single model, we restore corrupted images with different percentage of deleted pixels and denoise images of various noise levels.

\begin{figure*}
  \centering
  \includegraphics[scale=0.86]{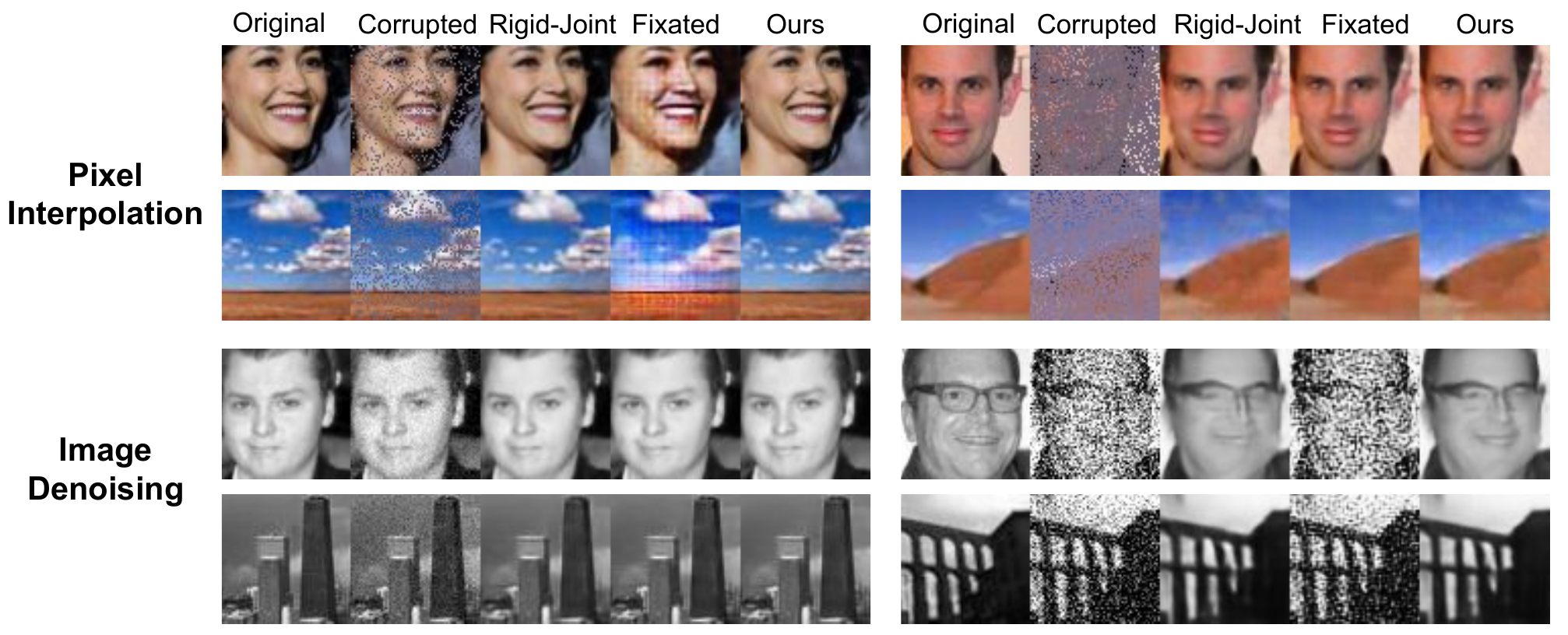}
  \caption{Qualitative examples of pixel interpolation and image denoising. For both tasks, the first row shows testing examples of CelebA dataset, and the second row shows examples of SUN397 dataset. For each quintuple, \textbf{Column 1}: Original image from the dataset; \textbf{Column 2}: Corrupted image; \textbf{Column 3}: Restored image using rigid joint training; \textbf{Column 4}: Restored image using a fixated model; \textbf{Column 5}: Restored image using our method. Models trained using our method can handle arbitrary levels of distortions, while the fixated models can only perform well at a particular level of corruption.}
  \label{fig:qualitative}
    \vspace{0.2in}
\end{figure*}

%===========================================================
\section{Image Denoising Results on DB11}

This section serves as a supplement to Section 6.7 in the main paper, where due to space constraints we could not describe the details of the setup of our image denoising system and present qualitative results. 

We first describe the details of our image denoising system. Because the input of our network is of size $64 \times 64$, given a larger corrupted image $\mathcal{C}$, we first decompose the image into overlapping patches of size $64 \times 64$ and use a sliding-window approach to denoise each patch separately (stride 3 pixels), \KG{then average outputs at overlapping pixels}.

We then present the qualitative results. Particularly,  we first compare the denoising results of image Lena across the spectrum of difficulty in Fig.~\ref{fig:lena}. We show image denoising results at four different corruption levels ($\sigma = 10, 25, 50, 75$). For each column, the first row shows the original real image; the second row shows the image corrupted by AWG noise with the specified sigma value; the third and fourth rows show the restoration results using KSVD~\cite{aharon2006img} and BM3D~\cite{dabov2007image} correspondingly assuming $\sigma=25$ for the test image; the fifth row shows the denoising result of the MLP~\cite{burger2012image} model trained for $\sigma=25$\footnote{We use the authors’ publicly available code (\url{http://people.tuebingen.mpg.de/burger/neural_denoising/}) in which the system is trained for $\sigma = 25$. The authors also propose a variant of the system trained on various corruption levels with $\sigma$ given as input to the network, and it requires the $\sigma$ value to be available at test time. This version is not available in the public code, and it is also unclear how the true $\sigma$ value would be available for a novel image with unknown distortions.}; the sixth row shows the restoration result using WCNN~\cite{gu2014weighted} assuming $\sigma=25$ for the test image; the seventh and eighth rows show the restoration results of the CSF~\cite{schmidt2014shrinkage} model and the TNRG~\cite{chen2016trainable} model trained for $\sigma=25$\footnote{We use the authors’ publicly available code (\url{https://github.com/uschmidt83/shrinkage-fields/}) and use the model trained for $\sigma = 25$.}\footnote{We use the authors’ publicly available code (\url{http://gpu4vision.icg.tugraz.at/index.php?content=downloads.php}) and use the model trained for $\sigma = 25$.} correspondingly; the last row shows the denoising result of the model trained using our on-demand learning algorithm. K-SVD, BM3D and WCNN only work well when given the correct sigma value at test time, which is impractical because it is difficult to gauge the corruption level in a novel image and decide which sigma value to use. The MLP, CSF, TNRG models trained for $\sigma=25$ are fixated models that perform well only at that specific level of corruption. However, the model trained using our proposed method performs well on all four corruption levels, and it is a single model without knowing the correct sigma value of corrupted images at test time. Finally, in the end we append the image denoising results using our denoising system of all the 11 images at noisy level $\sigma = 25$.

\onecolumn{

\begin{figure}
  \begin{center}
  \includegraphics[scale=0.45]{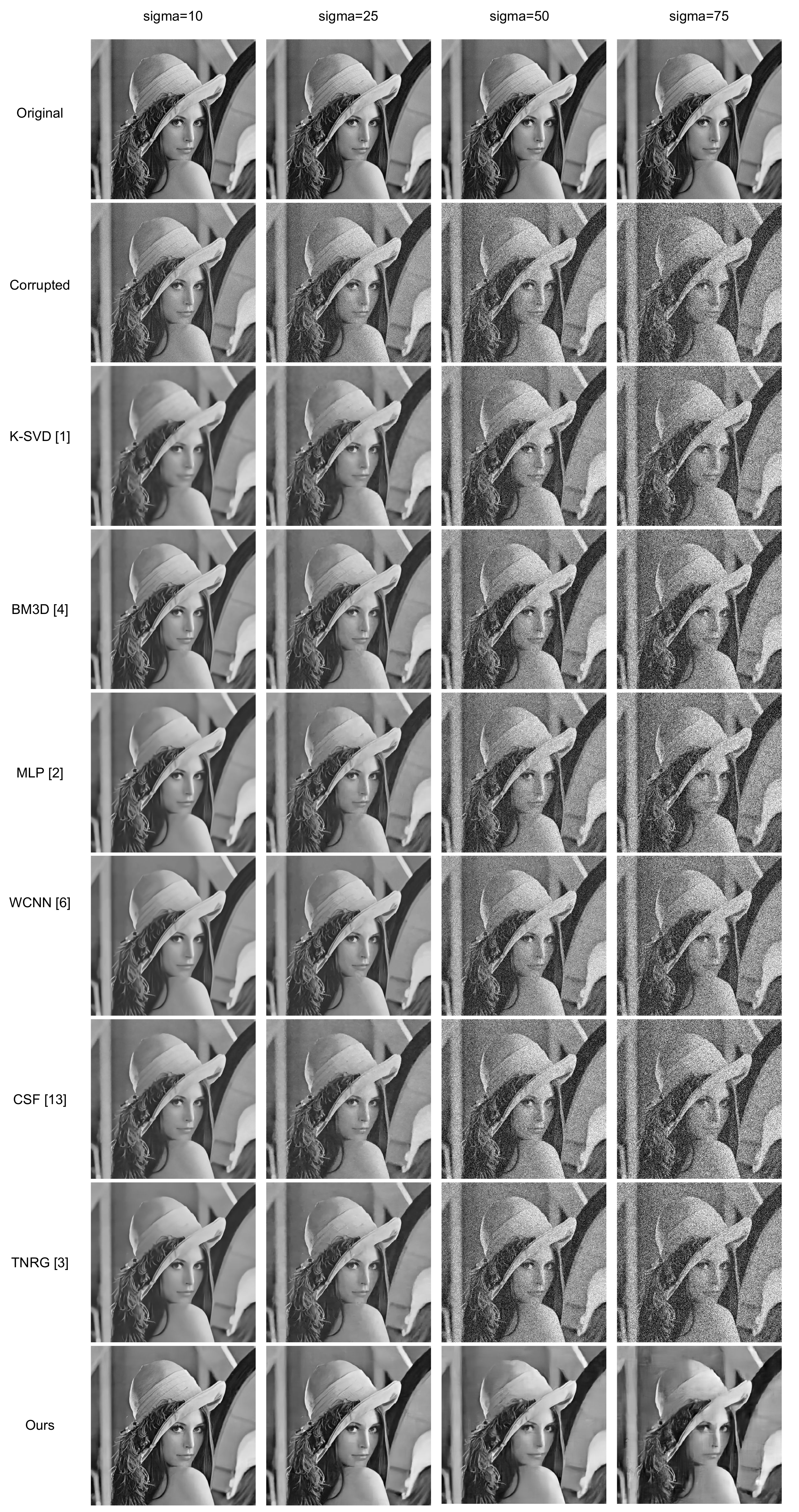}  	
  \end{center}
  \caption{Denoising results of image Lena at various corruption levels. All methods are applied as a single model to all test images. KSVD~\cite{aharon2006img}, BM3D~\cite{dabov2007image}, MLP~\cite{burger2012image}, WCNN~\cite{gu2014weighted}, CSF~\cite{schmidt2014shrinkage} and TNRG~\cite{chen2016trainable} perform well only at a particular level of corruption, while the image denoising model trained using our method performs well at all corruption levels.}
  \label{fig:lena}
\end{figure}
}

\twocolumn[{
\begin{center}
	\includegraphics[scale=0.54]{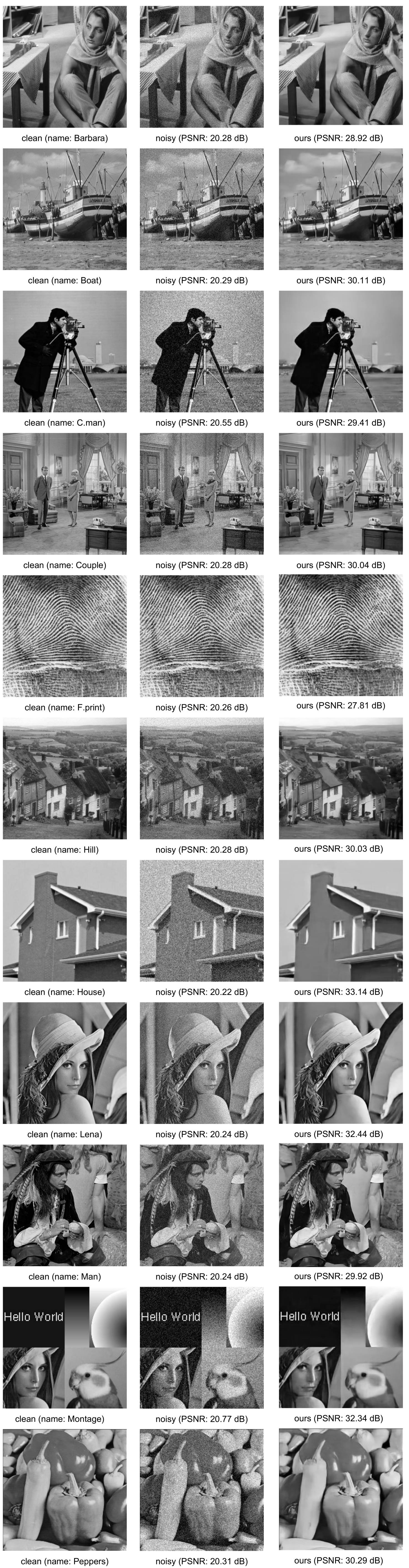}
\end{center}	
}]
\twocolumn[{
\begin{center}
	\includegraphics[scale=0.54]{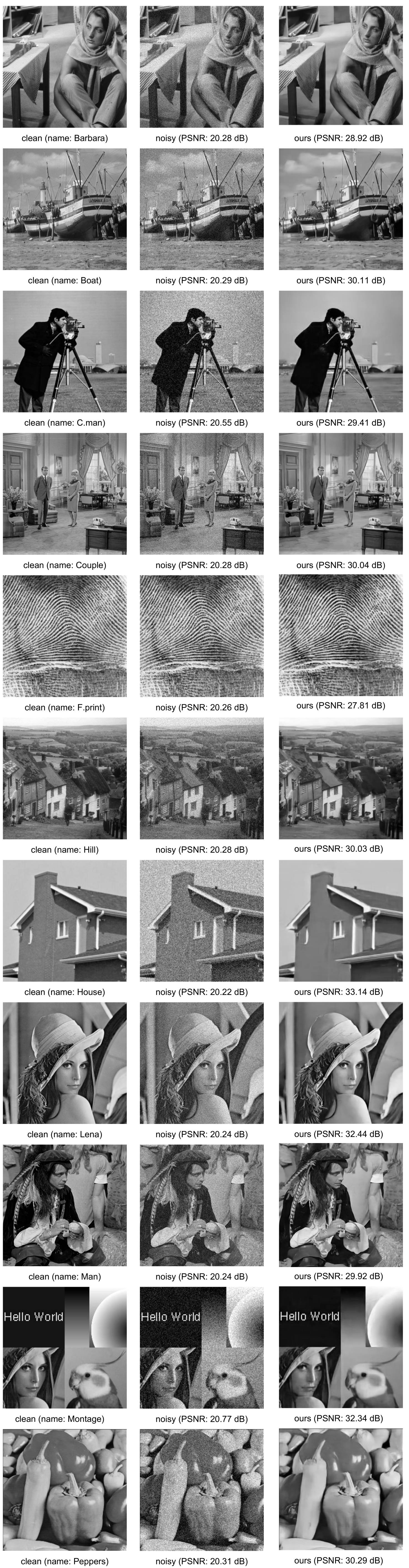}
\end{center}	
}]
\twocolumn[{
\begin{center}
	\includegraphics[scale=0.54]{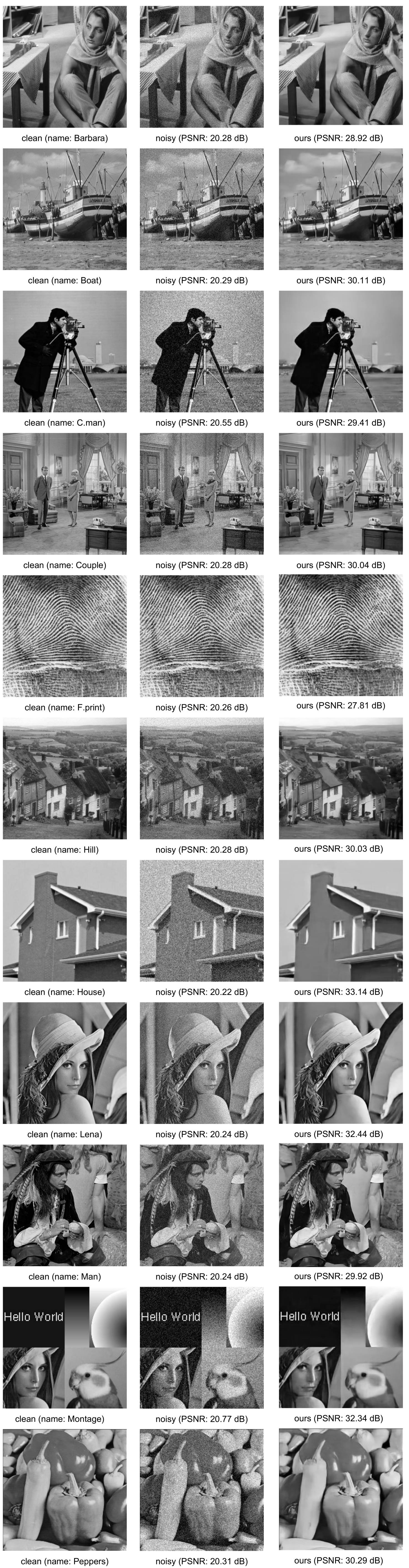}
\end{center}	
}]
\twocolumn
%======================================

\end{document}